\UseRawInputEncoding
\documentclass[preprint,12pt]{elsarticle}
\usepackage{amssymb}
\usepackage{placeins}
\usepackage{rotating}
\usepackage{colortbl}
\usepackage{lscape}
\usepackage{longtable}
\usepackage{adjustbox}
\usepackage{pdflscape}
\newcolumntype{L}{>{\centering\arraybackslash}m{6cm}}
\newcolumntype{M}{>{\centering\arraybackslash}m{2.5cm}l}

\biboptions{sort&compress}

\journal{Computer Methods and Programs in Biomedicine}

\begin{document}

\begin{frontmatter}

\title{Stabilizing Machine Learning for Reproducible and Explainable Results: A Novel Validation Approach to Subject-Specific Insights}
\author[inst1]{Gideon Vos}

\affiliation[inst1]{organization={College of Science and Engineering, James Cook University},
            addressline={James Cook Dr}, 
            city={Townsville},
            postcode={4811}, 
            state={QLD},
            country={Australia}}

\author[inst2]{Liza van Eijk}
\author[inst3]{Zoltan Sarnyai}
\author[inst1]{Mostafa Rahimi Azghadi}

\affiliation[inst2]{organization={College of Health Care Sciences, James Cook University},
            addressline={James Cook Dr}, 
            city={Townsville},
            postcode={4811}, 
            state={QLD},
            country={Australia}}

\affiliation[inst3]{organization={College of Public Health, Medical, and Vet Sciences, James Cook University},
            addressline={James Cook Dr}, 
            city={Townsville},
            postcode={4811}, 
            state={QLD},
            country={Australia}}
\begin{abstract}
\paragraph{Introduction}
\noindent Machine Learning (ML) is transforming medical research by enhancing diagnostic accuracy, predicting disease progression, and personalizing treatments. While general models trained on large datasets identify broad patterns across populations, the diversity of human biology, shaped by genetics, environment, and lifestyle, often limits their effectiveness. This has driven a shift towards subject-specific models that incorporate individual biological and clinical data for more precise predictions and personalized care. However, developing these models presents significant practical and financial challenges. To address this, this study introduces a novel validation approach that leverages a general ML model to achieve reproducible performance and robust feature importance analysis at both the group and subject-specific levels.

\paragraph{Methods}
\noindent We conducted initial experiments using a single Random Forest (RF) general model on nine datasets that varied in domain problems, sample size, and demographics. Different validation techniques were applied to assess model accuracy and reproducibility while evaluating feature importance consistency. Next, the experiment was repeated for each dataset for up to 400 trials per subject, randomly seeding the machine learning algorithm between each trial. This introduced variability in the initialization of model parameters, thus providing a more comprehensive evaluation of the machine learning model’s features and performance consistency. The repeated trials generated up to 400 feature sets per subject. The top subject-specific feature importance set across all trials was then identified. Finally, using all subject-specific feature sets, the top group-specific feature importance set was also created. 
The results of our proposed approach were then compared to the initial experiments using the conventional machine learning validation approaches in terms of performance and feature importance consistency.
\paragraph{Results}
\noindent We found that reproducibility, predictive accuracy, and feature importance were influenced by random seed selection and validation techniques during model training. Changes in random seeds or validation methods led to variations in feature importance and test accuracy, consistent with prior research on machine learning's sensitivity to random initialization. This study builds on that understanding by introducing a novel repeated trials validation approach with random seed variation, enabling consistent identification of key features for each subject using a single, generic machine learning model.

\paragraph{Conclusion}
Subject-specific models have been proposed as a solution to address the inherent variability in human biology and to achieve better model generalization. However, building and implementing these models in practice involves considerable cost and resource demands, making them potentially impractical for use in clinical trials. In this study, we introduce a novel validation technique for determining both group- and subject-specific feature importance within a general machine learning model, achieving higher predictive accuracy while providing feature importance consistency, which enhances model explainability.

\end{abstract}


\begin{highlights}
\item We show that generic machine learning model accuracy and feature importance are directly influenced by algorithm random seed selection, limiting results reproducibility and explainability.
\item We propose a novel validation approach using multiple randomized trials to stabilize model performance and feature importance for improved and consistent model explainability.
\item Our approach provides both group and subject-specific feature importance without loss of predictive power using a single generic machine learning model. 
\item In favor of reproducible research, the full source code, applied to 9 publicly available datasets, is made publicly available.
\end{highlights}


\end{frontmatter}


\section{Introduction and related work}

\noindent Machine learning (ML) and other forms of Artificial Intelligence (AI) have emerged as transformative statistical tools that are revolutionizing various fields of science, including medical research. These technologies excel at identifying complex patterns in vast datasets, often revealing insights that remain elusive to human researchers. In the realm of medicine, ML is increasingly utilized for tasks such as drug discovery \cite{Martinelli2022, Karampuri2024, Bhattacharjee2024}, patient diagnostics \cite{Zhu2023, Liu2024, Aslam2024}, and the analysis of genomic data \cite{Yuan2023a, Magazzu2022, Yang2022}. \\

\noindent However, the rapid integration of ML in research has also sparked concerns among scientists \cite{Ball2023, Kapoor2023, Ameli2024}, with fears that improper use is contributing to a surge in publications that make bold claims but lack the rigor necessary for reproducibility. Anecdotal evidence suggests that studies rife with errors and misleading results are becoming more common, raising questions about the reliability of findings in ML-driven research \cite{Ball2023, Kapoor2023, Ameli2024}. Although there has yet to be a systematic assessment of the scale of this issue, the potential consequences for medical science could be profound. \\

\noindent As researchers may increasingly rely on ML to guide critical decisions in healthcare, the need for robust validation, ethical and explainable application of these technologies becomes paramount to ensure that the advancements they offer are both meaningful and beneficial. A 2023 survey by Nature \cite{VanNoorden2023} found that while the use of ML is becoming increasingly common in science, 58\% of the 1,600 respondents raised concerns that ML techniques can introduce bias or discrimination in data, while 53\% noted that ill-considered use can lead to non-reproducible research. \\

\noindent Explainable A.I. (XAI) holds significant promise in making machine learning systems more transparent, trustworthy, and actionable \cite{Gunning2019, Rasheed2022}, potentially playing a crucial role in validating machine learning model predictions. Yet, the influence of ML recommendations on physician behavior remains poorly characterized. Nagendran \emph{et al.} \cite{Nagendran2023} investigated how clinicians' decisions may be influenced by additional information provided by XAI techniques, and found that information provided by ML and XAI had a strong influence on medical prescriptions. However, a key demand from clinicians, researchers and regulators alike is XAI which aims to not only provide recommendations but also justify or motivate the algorithmic reasoning to experts \cite{Nagendran2023}.\\

\noindent As machine learning models become more complex, the need for interpretability becomes vital in fields such as healthcare, where decisions based on these models can directly influence patient outcomes. To address this challenge, XAI supports two approaches to ensure explainability \cite{Arrieta2019}: \emph{(i)} ante-hoc explainability, which involves constructing models that are inherently transparent, and \emph{ii)} post-hoc explainability, which seeks to provide insights into complex, "black-box" models after they have been trained. Post-hoc techniques, such as feature importance analysis or visualization tools, aim to shed light on how these models arrived at their predictions, making them more interpretable without sacrificing performance. This study uses feature importance analysis for stabilizing machine learning model development.\\

\noindent A prerequisite of using XAI effectively for explainability is model generalization. Generalization refers to a model's ability to perform well on new, unseen data, ensuring that it captures the core patterns of the problem domain rather than over-fitting to the specifics of the training set. Without proper generalization, interpretability loses value and offers little meaningful insight. However, the notion of generalization itself may vary by context and demographics. A system that achieves the highest possible level of generalization is ideal, but an emphasis on overly broad generalization in healthcare applications may overlook specific scenarios where more context-specific machine learning models could offer better clinical utility \cite{Futoma2020}. \\

\noindent Several factors can impact a model's ability to generalize effectively across different contexts,  and collectively pose significant barriers to ensuring experimental results are reproducible. These include changes in clinical practice over time, patient demographic variation and changes to hardware and software used both for recording data and for model building and experimentation \cite{Futoma2020}. Additional considerations during model building are class imbalance, outlier management, bias mitigation and potential training data leakage, a flaw that occurs when information from the test or validation dataset is inadvertently included in the training dataset, leading to artificially inflated performance. \\

\noindent Kapoor \emph{et al.} \cite{Kapoor2023} surveyed a variety of medical research studies that utilized machine learning, and found 294 studies across 17 fields with potential data leakage, leading to overoptimistic findings. It is therefore crucial to first ensure that the machine learning model itself is robust, free from biases, and its results are reproducible across different contexts and datasets before utilizing XAI for interpretation, explanation, and validation against known clinical guidelines and standards. Without a solid foundation of robustness and reproducibility, the explanations provided by XAI may lack credibility, potentially leading to misguided decisions in sensitive areas such as clinical practice. \\

\noindent The primary objective of this study was to reproduce the findings of a previously published study \cite{Chekroud2024}, which provided source code, hardware and software specifications, and data via the Yale Open Data Access (YODA) Project. This effort aimed to validate the reproducibility and stability of machine learning results using a well-established dataset. Specifically, we focused on ensuring reproducible outcomes for both model performance and explainable feature importance.\\

\noindent Our analysis revealed that performance and feature importance reproducibility are highly sensitive to the selection of random seeds and validation techniques, which can undermine the reliability and generalization of ML models. To address this, we introduced a novel validation technique involving multiple random trials to stabilize both model performance and feature importance, enhancing the model's stability, as well as the reliability and interpretability of the results.

\section{Methods}

\subsection{Reproducibility} \label{experiment1}

\noindent The datasets used in this study are listed in Table \ref{tab:datasets}. For the first experiment, five international randomized controlled trial (RCT) datasets for evaluating the comparative efficacy of anti-psychotic medications for treating schizophrenia were utilized, as per the original study \cite{Chekroud2024} published in the journal \emph{Science}. These datasets are available from the YODA project as accession numbers NCT00518323, NCT00334126, NCT00085748, NCT00078039, and NCT00083668. \\

\noindent The aim of the first experiment was to reproduce the findings from the original study by re-running the pre-processing, analysis and visualization routines provided using the R source code made publicly available \cite{Chekroudcode}. Due to the resource intensive requirements of the original pre-processing and model building routines, only the Random Forest (RF) \cite{RANDOMFOREST} models were selected and run for a single set of outcome criteria, the Remission in Schizophrenia Working Group (RSWG).\\

\noindent Once the initial analysis was completed, all random seeds in the supplied source code were changed to a single number (42), and the process was re-run to test for results stability. We observed that changes in the random seed affected both feature importance and performance. To further evaluate our findings, we extended experimentation to additional well-studied and diverse public datasets (Table \ref{tab:datasets}, 2-8).

\begin{table}[!t]
\centering
\caption{\label{tab:datasets}Datasets utilized in this study.}
\resizebox{\textwidth}{!}{
\begin{tabular}{lcccc}
\hline\hline
\textbf{Dataset} & \textbf{Sample Size} & \textbf{Features} & \textbf{Ordinals} & \textbf{Cardinality}  \\
\hline
1. YODA RCT \cite{Chekroud2024} & 1513 & - & - & - \\ 
\rowcolor[rgb]{0.753,0.753,0.753} 2. Breast Cancer \cite{datasetcancer}    & 683                  & 10                & 10                & 91                    \\
3. Diabetes \cite{datasetdiabetes}         & 351                  & 35                & 3                 & 8150                  \\
\rowcolor[rgb]{0.753,0.753,0.753} 4. College \cite{datasetcollege}          & 777                  & 18                & 1                 & 6249                  \\
5. Cars \cite{datasetcars}             & 32                   & 11                & 5                 & 171                   \\
\rowcolor[rgb]{0.753,0.753,0.753} 6. Glaucoma \cite{datasetglaucoma}        & 196                  & 63                & 1                 & 8960                  \\
7. Glass \cite{datasetglass}            & 214                  & 10                & 1                 & 945                   \\
\rowcolor[rgb]{0.753,0.753,0.753} 8. Diamonds \cite{datasetdiamonds}         & 250                  & 10                & 3                 & 544                   \\
8. Diamonds \cite{datasetdiamonds}         & 500                  & 10                & 3                 & 737                   \\
\rowcolor[rgb]{0.753,0.753,0.753} 8. Diamonds \cite{datasetdiamonds}         & 2000                 & 10                & 3                 & 1273                  \\
8. Diamonds \cite{datasetdiamonds}         & 5000                 & 10                & 3                 & 2077                  \\
\rowcolor[rgb]{0.753,0.753,0.753} 9. Alzheimer's Disease \cite{datasetalzheimers}         & 48                   & 23                & -                 & 448                   \\
\hline
\end{tabular}
}
\end{table}
\FloatBarrier

\subsection{Effect of random seed and validation techniques on performance and feature importance } \label{experiment2}

\noindent The next set of experiments utilized the RF algorithm in order to build models for predicting the relevant binary outcome labeled within each dataset. These experiments were designed to evaluate how altering random seeds during the initialization of the RF algorithm influences accuracy metrics and the variance in feature importance reported by the model. Furthermore, the effect of different validation techniques on model performance and feature importance was examined. \\

\noindent The datasets shown in Table \ref{tab:datasets} were selected due to the varying sample sizes, feature attributes, and accessibility as open datasets. Table \ref{tab:datasets}  further details the number of features, ordinal variables (categorical variables with a meaningful order), and cardinality (the number of unique values a categorical variable can take) for each of the nine datasets. These attributes were included in the table to highlight that their counts had no observable correlation with the stability of feature importance during subsequent experiments and provided sufficient variation for results comparison. All experimentation was done using R \cite{RCite} version 4.4.1. \\

\noindent For each experiment and related dataset, a number of validation techniques were applied including an 80\%/20\% train and test split, leave-one-subject-out (LOSO) validation, 10-fold cross validation, and leave-one-out cross-validation (LOOCV), with each training and validation round repeated using two different random seeds to initialize the RF algorithm (42 and 43). Feature importance was determined using the built-in importance methods provided using RF, along with Shapley values (SHAP) \cite{SHAP} and Local Interpretable Model-agnostic Explanations (LIME) \cite{LIME}.\\

\subsection{Proposed Randomized Trials Validation Approach} \label{experiment3}

\noindent The RF algorithm uses a user-specified random seed during bootstrap sampling and random feature selection to create multiple subsets of the training data \cite{Breiman2001} and ensure reproducibility between model training sessions \cite{Breiman2001, Beam2020}. However, a study by Henderson \emph{et al.} \cite{Henderson2018} found that altering the random seed could inflate the estimated model performance by as much as 2-fold, relative to what a different set of random seeds would yield. Additionally, Peng \emph{et al.} \cite{Peng2011} noted that system-specific factors including software library versions and hardware specifications can influence the consistency of results when machine learning models are re-run, by potentially impacting the underlying random number generator.  \\

\begin{figure}[h!]
\centering
\includegraphics[width=\textwidth]{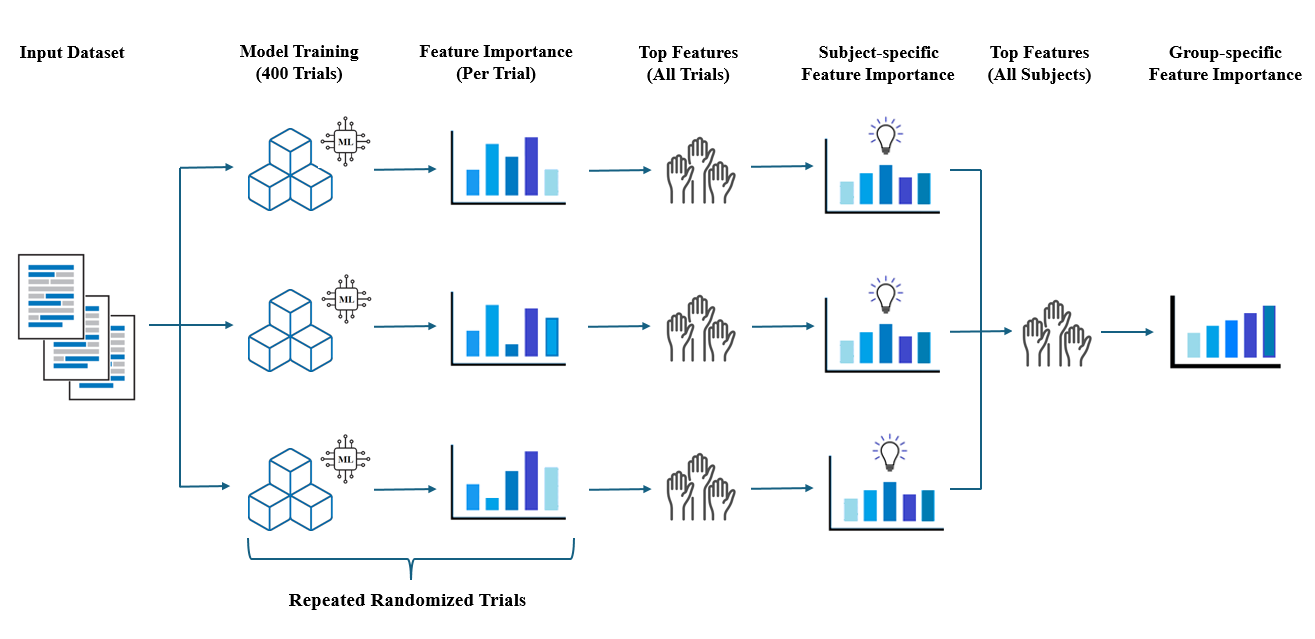}
\caption{\label{fig:figure1} Proposed randomized trial validation approach for subject- and group-specific feature importance and model performance stabilization.}
\end{figure}

\noindent In order to compensate for these noted limitations related to random seed selection and its potential impact on reproducibility, we addressed the problem through a repeated randomized validation method as detailed in Figure \ref{fig:figure1}.

\FloatBarrier

\noindent The proposed method first splits the training dataset by subject (see the first column in Fig. \ref{fig:figure1}). For each subject, a random seed is created using the subject index and trial number. A model is then trained on data from all other subjects and tested on the current subject (LOSO validation), and repeated for up to 400 trials. Note that experimentation using trial counts ranging from 50 to 1,000 showed an optimal maximum of 400. If the model correctly predicts the outcome in each trial, the most important features are recorded. After completing all 400 trials for a subject (column 2 in Fig. \ref{fig:figure1}), the recorded feature importance sets across the 400 trials (column 3 in Fig. \ref{fig:figure1}) are grouped and ranked using voting (column 4 in Fig. \ref{fig:figure1}) to identify the top sets that contributed most to achieving 100\% accuracy, per subject (column 5 in Fig. \ref{fig:figure1}). Upon completion of all subjects across all trials, the same ranking method (column 6 in Fig. \ref{fig:figure1}) is applied to find the feature sets that occur most often across the dataset. This specifies the overall feature importance for the group (column 7 in Fig. \ref{fig:figure1}). \\

\noindent The proposed method was tested with datasets 1 to 8 (Table \ref{tab:datasets}) to confirm its effectiveness across diverse domains, beyond just medical datasets. Additionally, for dataset 8, multiple sample sizes were selected and run to compare results on a single dataset of varying sizes. \\

\noindent Finally, to validate the use of the proposed method to achieve stable performance metrics and feature importance (per subject and per group) in a medical dataset, the dataset from \cite{datasetalzheimers} was utilized. This dataset 9 consists of 34 healthy controls and 14 subjects identified as having Alzheimer's disease, and was previously analyzed \cite{Besga2015} to assess the relative significance of clinical observations, neuropsychological tests, and specific blood plasma biomarkers (inflammatory and neurotrophic) \cite{datasetalzheimers}.  \\

\section{Results}
\subsection{Challenges in reproducibility}
\begin{figure}[h!]
\centering
\includegraphics[width=\textwidth]{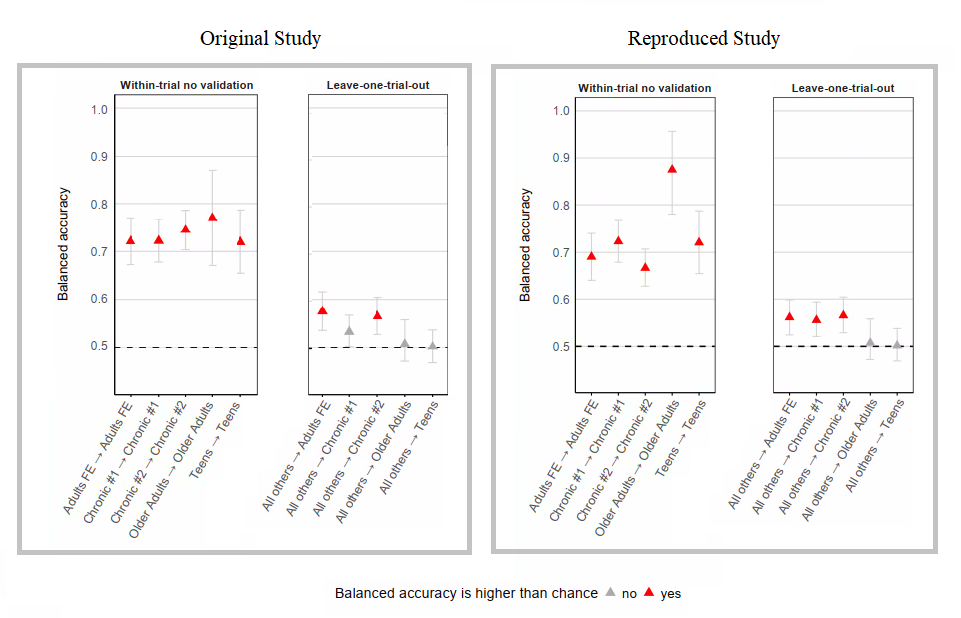}
\caption{\label{fig:figure2}Original published results from \cite{Chekroud2024} vs.~reproduced results with a different random seed using the published source code \cite{Chekroudcode}.}
\end{figure}

\noindent Figure \ref{fig:figure2} shows a comparison between the results of two scenarios (Within-trial no validation and Leave-one-trial-out) reported in the original study by Chekroud \emph{et al.} \cite{Chekroud2024}, and those found in our reproduction (as explained in section \ref{experiment1}) after adjusting the original random seed numbers in the published source code \cite{Chekroudcode}. For the within-trial no validation scenario, we note a substantial difference in both Chronic \#2 and Older Adult subsets, with less substantial differences in the Leave-one-trial-out scenario. However, balanced accuracy and classification quality metrics reported across both scenarios showed virtually no difference between the original study (0.737, 0.537) and our reproduced study (0.735, 0.539). \\

\noindent Further adjusting the random seed again showed differences within the individual scenarios. These findings are consistent with those of the original study, and further underscore the challenges in reproducing study results, even under ideal conditions where the source code, data, and hardware platforms have been duplicated. Based on the results of the original study, the authors suggested that machine learning models predicting treatment outcomes (in schizophrenia) are highly context-dependent and may have limited generalizability. A potential approach to address variations in experimental context and patient demographics is the development of patient-specific models \cite{Chen2017, Kopitar2020, Goldstein2016, Bzdok2017, Obermeyer2016}. However, this solution introduces greater complexity, logistical challenges, and increased costs related to real-world implementation and scaling. Therefore, a new general method is needed to improve reproducibility and generalizability across diverse contexts, and this is the focus of this work.\\

\FloatBarrier

\noindent Bouthillier \emph{et al.} \cite{Bouthillier2019} identified three types of study reproducibility: 
\begin{itemize}
\item Methods Reproducibility: A method is reproducible if reusing the original code leads to the same results.
\item Results Reproducibility: A result is reproducible if a re-implementation of the method generates statistically similar values.
\item Inferential Reproducibility: A finding or a conclusion is reproducible if one can draw it from a different experimental setup.
\end{itemize}

\noindent Based on the above criteria, our initial experiment using the code and data \cite{Chekroudcode} from \cite{Chekroud2024} fell short in all three aspects when considering not only overall balanced accuracy and classification quality metrics, but also specific scenarios within the study (Figure \ref{fig:figure2}). This limitation becomes particularly relevant when feature importance ranking, alongside interpretability, is a primary study outcome \cite{CiobanuCaraus2024}, beyond mere predictive performance. \\

\noindent Breiman \cite{Breiman2001} initially introduced an ad hoc, computationally efficient feature importance calculation method for the RF algorithm known as "out-of-bag" variable importance (OOB VIMP) which remains the default in most implementations and is widely used in the research community, despite significant limitations. Wallace \emph{et al.} \cite{Wallace2023} evaluated OOB VIMP's and proposed "knockoff VIMPs", an improved method which facilitates a direct and interpretable estimate of the value of a feature within a model, while however, still resulting in individual variables that can be challenging to interpret.\\

\noindent Henderson \emph{et al.} \cite{Henderson2018} in their study on RF feature importance metrics in medicine, suggested the use of proper significance testing and multiple trials with varying random seeds when comparing predictive performance, with random seed selection explicitly part of the algorithm. This averaging of multiple runs over different random seeds can give insight into the population distribution of the algorithm performance in an environment \cite{Henderson2018}. Building upon this and the aforementioned works, we developed a new validation approach aimed at stabilizing reproducibility, enhancing generalization, and improving the explainability of machine learning models.\\

\subsection{Evaluating random trial validation}
\noindent To thoroughly assess the effectiveness of our proposed randomized validation approach, we conducted experiments using a selection of well-researched open datasets (Table \ref{tab:datasets}, datasets 2 - 8) as detailed in sections \ref{experiment2} and \ref{experiment3}. Among these seven datasets, three are related to health care, whereas four non-health care datasets were chosen to eliminate any potential concerns related to population diversity. Below, we discuss the results from all health-related datasets. \\

\begin{figure}[!t]
\centering
\includegraphics[width=\textwidth]{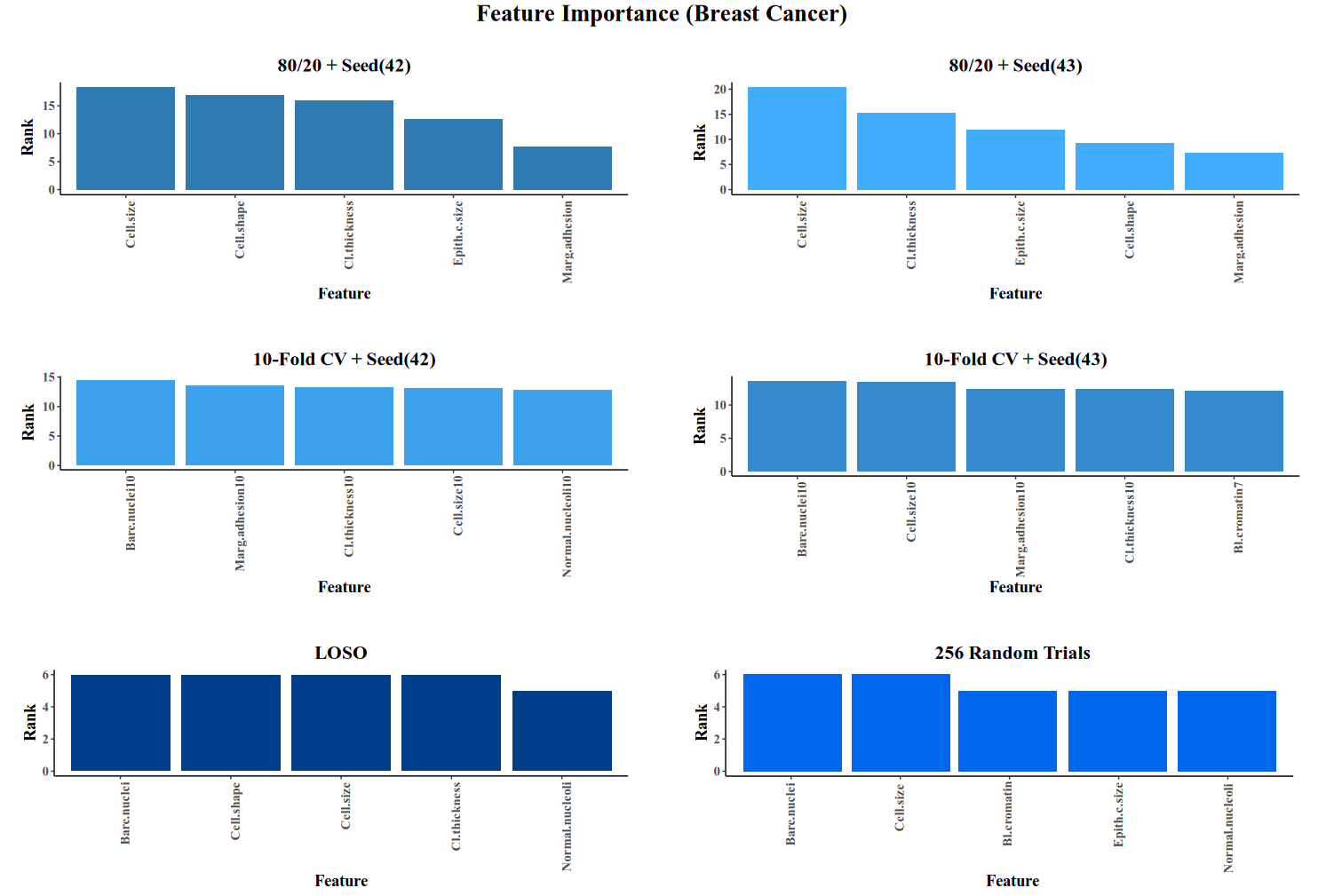}
\caption{\label{fig:figure3}Experimental results on the Breast Cancer dataset \cite{datasetcancer}. The figure shows how modifying the cross-validation technique and/or random seed can result in different feature importance sets, undermining model generalization, stability, and explainability. The figure also shows a stabilized feature importance set, using our proposed random trial validation technique.}
\end{figure}

\noindent Figure \ref{fig:figure3} details the outcome of experimentation using RF on the breast cancer dataset (Table \ref{tab:datasets}, \#2) \cite{datasetcancer}. Where a 80\%/20\% train/test split method was used for validation, the cell size feature consistently ranked as the most important, irrespective of random seed choice. However, feature ranking for the next four most important features differed significantly  (see the first row of Fig. \ref{fig:figure3}). The same effect can be observed when other validation methods were utilized (10-Fold CV, LOSO) with varying random seeds, where bare nuclei ranked as the top most important feature. Using the proposed random trial validation method described in section \ref{experiment3}, we reach feature importance stability within 256 trial iterations across all subjects as a group. Additionally, we reach a stable feature importance set per subject (not shown) using our proposed technique shown in Fig. \ref{fig:figure1}. \\

\FloatBarrier

\begin{figure}[!t]
\centering
\includegraphics[width=\textwidth]{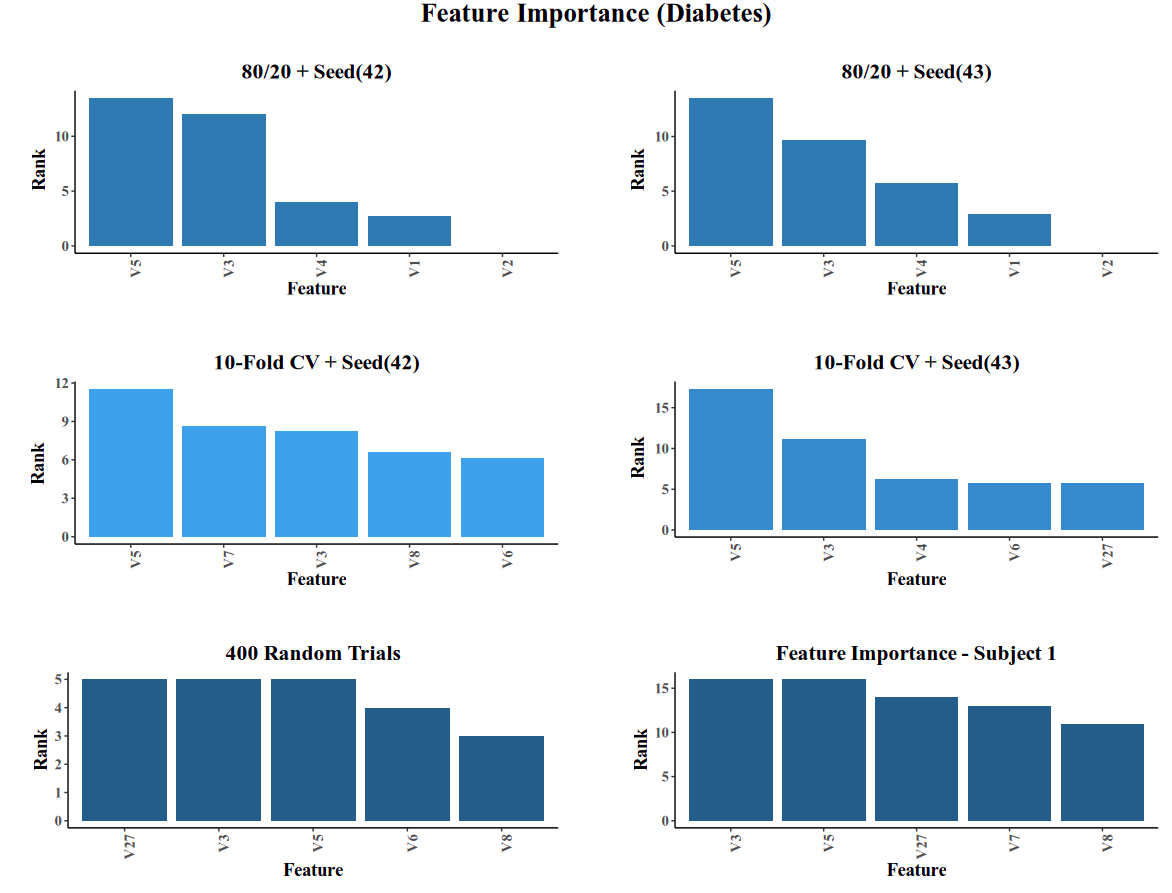}
\caption{\label{fig:figure4}Experimental results on the Diabetes dataset \cite{datasetdiabetes}. The figure shows how modifying the cross-validation technique and/or random seed can result in different feature importance sets, undermining model generalization, stability, and explainability. The figure additionally presents a stabilized feature importance set for the entire dataset subjects (third row, left column) and for a sample individual subject 1 (right column), employing our proposed validation method of random trials.}
\end{figure}

\noindent Experimentation on the diabetes dataset \cite{datasetdiabetes} (Figure \ref{fig:figure4}) yielded matching feature importance ranks irrespective of random seed choice when using a 80\%/20\% train/test split for validation (see the first row of Fig. \ref{fig:figure4}). However, the results are markedly different compared to those of the 10-Fold CV method (second row of Fig. \ref{fig:figure4}). Additionally, there is significant variation in feature importance rank using two different seed values, further demonstrating the instability in machine learning on this dataset. By applying the proposed random trial validation method, we reached feature importance stability within 400 trials. Importantly, subject-level feature importance (for Subject 1 as a sample) differs from overall group feature importance.\\

\noindent Similar observations were noted for experiments performed on datasets 4 to 8 (Table \ref{tab:datasets}), irrespective of sample size, which was tested by using dataset 8 (diamond classification). Because of space constraints, these findings are not included in the main text but can be reviewed and replicated through our open-source code and data available at https://github.com/xalentis/Reproducibility.\\ 

\FloatBarrier

\begin{figure}[!t]
\centering
\includegraphics[width=\textwidth]{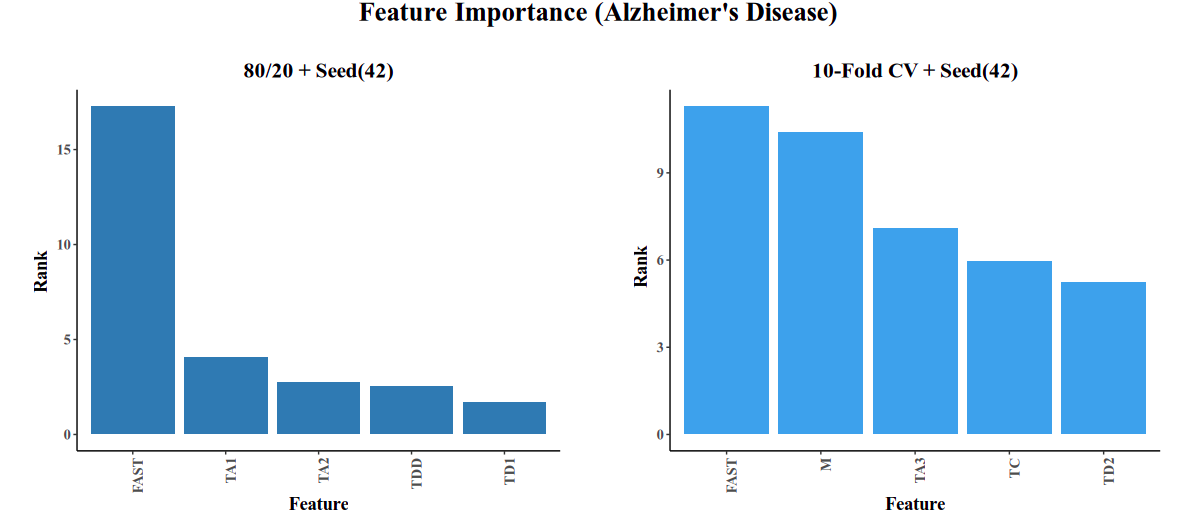}
\caption{\label{fig:figure5} Experimental results on the Alzheimer's disease dataset \cite{datasetalzheimers}. The figure shows how modifying the cross-validation technique even when the random seed is kept the same results in different feature importance sets, undermining model generalization, stability, and explainability.}
\end{figure}

\noindent Dataset 9 (Alzheimer's disease) \cite{datasetalzheimers} was used to validate the proposed random trial approach in a practical, real-world scenario. Figure \ref{fig:figure5} shows the feature importance plots generated using the default OOB VIMP approach implemented in the RF algorithm. When comparing the outcomes from an 80\%/20\% train/test split to those from 10-Fold CV, we observe variations in feature importance rankings, including four completely distinct features among the top five, aside from FAST, which is ranked as the most critical in both cases. This highlights how easily a model's explainability and stability can be influenced by merely choosing a different validation method. \\

\FloatBarrier

\subsection{Comparative analysis and validation insights into feature importance}
\noindent In a prior study by Besga \emph{et al.} \cite{Besga2015} utilizing Support Vector Machines (SVM), CART Decision Trees, and RF to classify healthy controls from subjects diagnosed with Alzheimer's disease using dataset 9, a Welch's t-test p-value for each behavioral, biological biomarker, and the aggregate neuropsychological feature was calculated. Results from this study indicated FAST, apathy (TA3), executive functions (EF), attention (A) and memory (M) as the most statistically significant features ($p < 0.001$) followed by total sleep (TS), total anxiety (TA2) and total dysphoria/depression (TDD). \\

\noindent Based on the results from \cite{Besga2015}, figure \ref{fig:figure5} prompts an inquiry into which of the two validation strategies better highlights the most relevant top-tier features. The 80\%/20\% validation scheme ranks the FAST feature among its top five features with $p < 0.001$; however, it also includes TA1 and TD1, which are absent from the top eight statistically significant features identified in \cite{Besga2015}. In contrast, the 10-fold CV approach includes three features in its top five that are statistically significant at $p < 0.001$, while the other two features (TC and TD2) do not appear in the top eight significant features according to the same study.\\

\begin{figure}[!t]
\centering
\includegraphics[width=\textwidth]{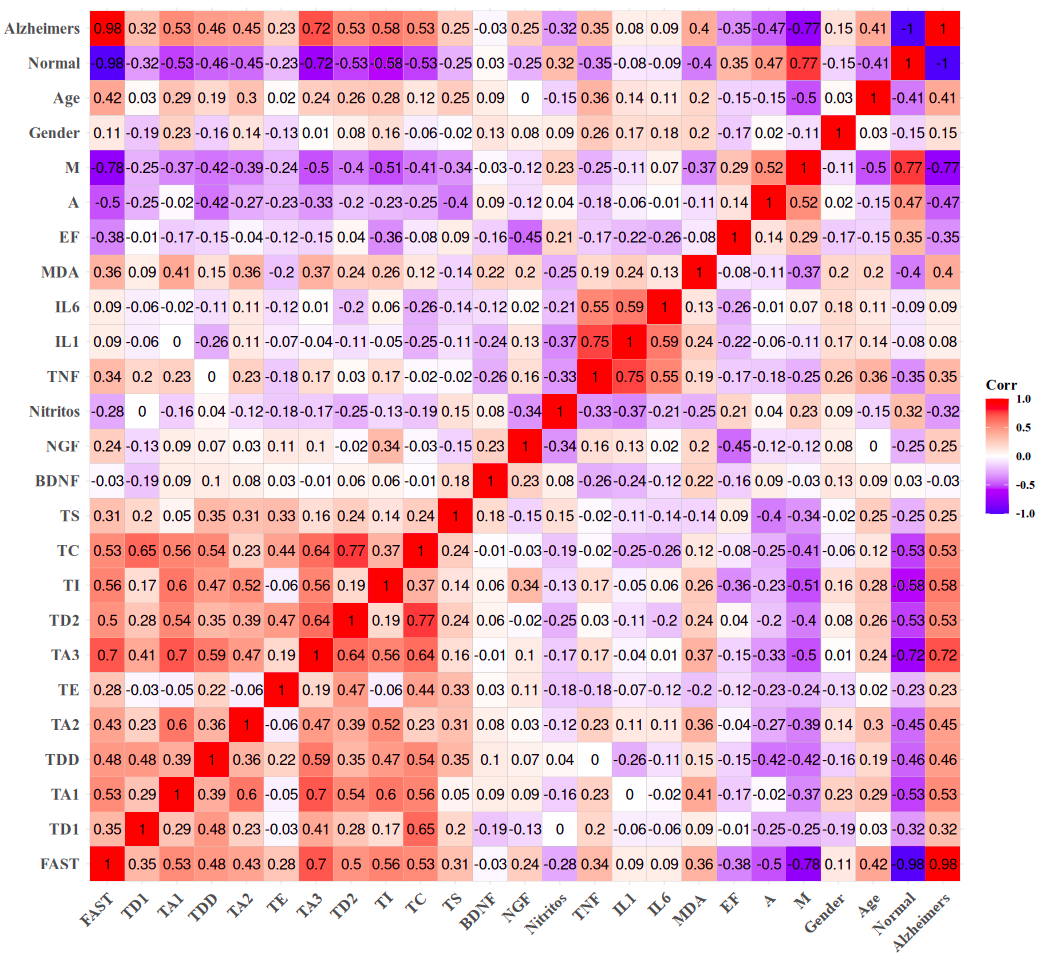}
\caption{\label{fig:figure6} Spearman correlations between 23 features and two classes (i.e. Normal and Alzheimers) within dataset \cite{datasetalzheimers}.}
\end{figure}
\FloatBarrier

\noindent To assess the significance of features in the Alzheimer's dataset \cite{datasetalzheimers}, Spearman correlations were computed between the 23 features and the two classes (Normal and Alzheimer's), as depicted in Fig. \ref{fig:figure6}. This figure is consistent with the results of \cite{Besga2015} and offers an additional approach to corroborate our proposed randomized trial method.\\

\begin{figure}[!t]
\centering
\includegraphics[width=\textwidth]{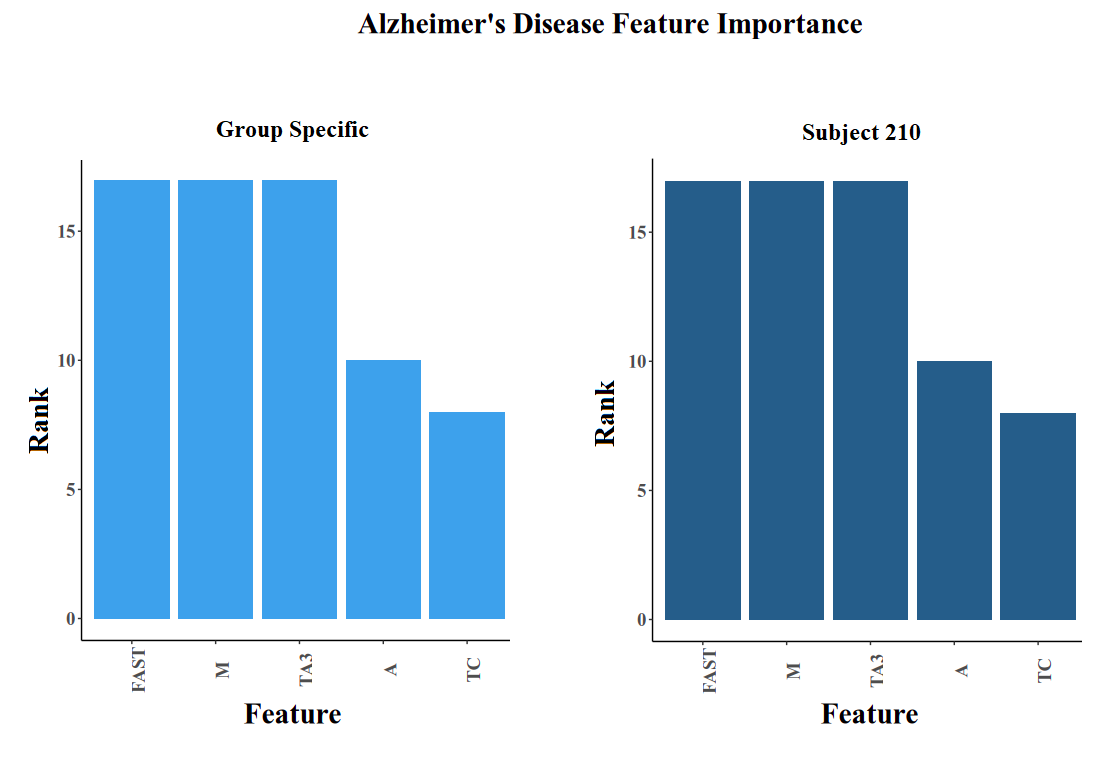}
\caption{\label{fig:figure7} Experimental results on the Alzheimer's disease dataset \cite{datasetalzheimers} using the proposed randomized trial method.}
\end{figure}

\noindent Figure \ref{fig:figure7} shows the feature importance obtained using the proposed random trial method for the group (left) and for an individual subject (right). The feature rankings demonstrate a strong correlation with previous research \cite{Besga2015} and the findings shown in  Fig. \ref{fig:figure6}, particularly for FAST, M, TA3 and A. In contrast to the outcomes shown in Figure \ref{fig:figure5}, our proposed method successfully identifies four statistically significant features with $p < 0.001$ among its top-5 rankings, assigning them all high ranks. This result is not achieved in any of the experiments depicted in Figure \ref{fig:figure5}. 
Additionally, our proposed method yields stable group and subject-level feature importance that correlates well with prior clinical findings on biomarkers significant in Alzheimer's disease \cite{Besga2015}, irrespective of random seed choice for algorithm initialization. For this particular dataset, all individual subject-level feature importance ranks corresponded with those at the group level.\\

\FloatBarrier

\subsection{Comparative evaluation of validation techniques: accuracy and computational efficiency}
\noindent Table \ref{tab:accuracy} provides a comparison of accuracy scores achieved for two sample datasets included in this study (the Breast Cancer \cite{datasetcancer} and the Alzheimer's Disease \cite{datasetalzheimers}), using a variety of common validation methods, along with  our proposed validation method using random trials. Our proposed approach attained similar accuracy levels on both datasets when matched against the three traditional validation techniques. Nevertheless, Table \ref{tab:accuracy} illustrates the impact of the chosen validation technique on model performance. It is important to highlight that, given the limited sample size of 48 in the Alzheimer's Disease dataset \cite{datasetalzheimers}, only LOSO and our proposed validation methods were employed. \\

\begin{table}[!t]
\centering
\caption{\label{tab:accuracy}Validation method accuracy comparison.}
\resizebox{\textwidth}{!}{
\begin{tabular}{lclc}
\hline\hline
\multicolumn{1}{c}{\textbf{Dataset}} & \multicolumn{1}{c}{\textbf{Sample Size}} & \multicolumn{1}{c}{\textbf{Validation}} & \multicolumn{1}{c}{\textbf{Accuracy}}  \\
\hline\hline
2. Breast Cancer \cite{datasetcancer}                     & 630                                      & 80/20                                   & 100.00\%                               \\
\rowcolor[rgb]{0.753,0.753,0.753} 2. Breast Cancer \cite{datasetcancer}                     & 630                                      & 10-Fold CV                              & 97.00\%                                \\
2. Breast Cancer \cite{datasetcancer}                     & 630                                      & LOOCV                                   & 97.00\%                                \\
\rowcolor[rgb]{0.753,0.753,0.753} 2. Breast Cancer \cite{datasetcancer}                     & 630                                      & Random Trials                           & \textbf{99.50\%}                       \\
9. Alzheimer's Disease \cite{datasetalzheimers}                           & 48                                       & LOSO                                    & 100.00\%                                \\
\rowcolor[rgb]{0.753,0.753,0.753} 9. Alzheimer's Disease \cite{datasetalzheimers}                           & 48                                       & Random Trials                           & \textbf{100.00\%}                       \\
\hline
\end{tabular}
}
\end{table}
\FloatBarrier

\noindent Table \ref{tab:performance} provides a summary of the execution times (in minutes) for each validation experiment conducted on datasets 2-8 in this study. Over 400 randomized trials, our proposed method significantly reduced the computational time required compared to the standard LOSO approach, while taking longer time than the 10-fold cross-validation (CV). The 80\%/20\% method resulted in much shorter experimentation times. Although our approach has a longer execution time compared to 10-fold CV and 80\%/20\% methods, it offers the advantages of yielding stable, reproducible accuracy scores and reliable feature importance assessments on both a group and subject level.

\section{Conclusion and Discussion}

\noindent In this study, we introduced a novel validation technique for determining and stabilizing both group-level and subject-specific feature importance within a single, generalised machine learning framework. This approach addresses the inherent variability in human biology, a key factor that often complicates the reproducibility and interpretability of machine learning results, even when the same hardware and software settings are applied. We performed an array of experiments on several open datasets to evaluate the performance of our approach.\\

\noindent Using a sample dataset, we demonstrated that the proposed validation method achieves high Spearman correlation levels between expected and predicted feature importance, aligning with established biomarkers for Alzheimer's disease, thereby underscoring its clinical relevance. Moreover, we showed that the method delivers predictive accuracy and runtime performance comparable to widely used validation techniques, offering a stable and interpretable alternative for biomedical applications.\\

\begin{table}[!t]
\centering
\caption{\label{tab:performance}Validation method execution time comparison.}
\resizebox{\textwidth}{!}{
\begin{tabular}{lccccc}
\hline\hline
\multicolumn{1}{c}{\textbf{Dataset}} & \multicolumn{1}{c}{\textbf{Sample Size}} & \multicolumn{1}{c}{\textbf{400 Random Trials (mins.)}} & \multicolumn{1}{c}{\textbf{LOSO (mins.)}} & \multicolumn{1}{c}{\textbf{10-Fold CV (mins.)}} &  {\textbf{80/20 (secs)}}  \\
\hline
2. Breast Cancer \cite{datasetcancer}                     & 683                                      & 7.2                                     & 15                                & 4   &      0.1                                \\
\rowcolor[rgb]{0.753,0.753,0.753} 3. Diabetes \cite{datasetdiabetes}                          & 351                                      & 1.6                                     & 2                                 & 1 &       0.2                                 \\
4. College \cite{datasetcollege}                           & 777                                      & 2.4                                     & 7                                 & 2      &   0.6                                \\
\rowcolor[rgb]{0.753,0.753,0.753} 5. Cars \cite{datasetcars}                              & 32                                       & 0.07                                    & 0.38                              & 0.05 &      0.01                                \\
6. Glaucoma \cite{datasetglaucoma}                          & 196                                      & 1.2                                     & 1                                 & 1       &   0.2                               \\
\rowcolor[rgb]{0.753,0.753,0.753} 7. Glass \cite{datasetglass}                             & 214                                      & 0.6                                     & 0.42                              & 0.4 & 0.1                                      \\
8. Diamonds \cite{datasetdiamonds}                          & 250                                      & 2                                       & 2                                 & 1     & 0.3                                   \\
\rowcolor[rgb]{0.753,0.753,0.753} 8. Diamonds \cite{datasetdiamonds}                          & 500                                      & 4                                       & 7                                 & 2  & 0.6                                       \\
8. Diamonds \cite{datasetdiamonds}                          & 2000                                     & 16                                      & 99                                & 11     & 3                                  \\
\rowcolor[rgb]{0.753,0.753,0.753} 8. Diamonds \cite{datasetdiamonds}                          & 5000                                     & 38.4                                    & 600                               & 29                & 7       \\
9. Alzheimer’s Disease \cite{datasetalzheimers} & 48 & 
 6.1 & 0.14 & 0.064 & 0.02 
\\
\hline
\end{tabular}
}
\end{table}
\FloatBarrier

\noindent A limitation of our proposed approach is its higher computational demand compared to widely-used techniques such as 10-fold cross-validation and the 80/20 validation split. However, our method demonstrated significantly improved computational efficiency compared to the commonly used LOSO technique, which is prevalent in medical machine learning research. Despite its increased computational cost, the enhanced stability in reproducibility and explainability offered by our approach provides a valuable trade-off, making it a worthwhile option in medical AI, where these factors are critical.\\

\noindent Additionally, the proposed method represents a significant advancement in the pursuit of explainable AI by reliably identifying stable feature importance. By tailoring feature importance to both individual patients and broader subject cohorts, this approach enhances the interpretability of model decisions while providing medical practitioners with clearer insights into the rationale behind the model's recommendations. This insight can help identify more cost-effective group-level diagnostic, prognostic, and predictive features, streamlining the process of biomarker selection. By prioritising features based on factors such as cost, risk, and invasiveness, this approach can optimise resource allocation and improve clinical outcomes. At the subject level, the feature importance can guide the collection of only the most relevant and significant biomarkers for each individual, enhancing patient outcomes and reducing unnecessary clinical costs.\\

\noindent While innovations in medical machine learning, like the proposed approach, have the potential to make a significant impact, their effectiveness will be limited if the study findings are not reproducible and accessible for further research. Open access to data and code is vital for advancing scientific research, particularly in health care, where reproducibility and transparency are essential for fostering trust in AI-driven solutions. By supporting the replication of results and encouraging further exploration, open access facilitates the development of robust, generalisable, and clinically impactful AI models, which is the primary aim of our study. To support further research in this area, the full R source code used in this study is available on GitHub at https://github.com/xalentis/Reproducibility.

 \bibliographystyle{elsarticle-num} 
 \bibliography{cas-refs}






\end{document}